\documentclass[a4paper,11pt]{article}
\usepackage{threeparttable}
\usepackage[left=2cm, right=2cm, top=2cm, bottom=2cm]{geometry}
\usepackage{amsmath,amsfonts,amssymb} 
\usepackage{setspace}            

\usepackage[blocks]{authblk}

\usepackage{graphicx}                 
\usepackage{subcaption}               
\usepackage{lscape}                   
\usepackage{pdfpages}                
\usepackage[dvipsnames]{xcolor}      

\usepackage{booktabs}                
\usepackage{longtable}                
\usepackage{tabularx}                 
\usepackage{multirow}                
\usepackage{makecell}                
\usepackage{array}                    
\usepackage{float}                    
\usepackage{enumitem}                
\usepackage{multicol}                 
\usepackage{tocloft}                  
\usepackage[semicolon]{natbib}        
\usepackage{hyperref}                

\hypersetup{
    colorlinks=true,
    linkcolor=black,
    citecolor=black,
    filecolor=black,      
    urlcolor=black,
}
\title{\bfseries\Large Grounding Large Language Models in Clinical Evidence: A Retrieval-Augmented Generation System for Querying UK NICE Clinical Guidelines}

\author[1]{Matthew Lewis\thanks{Corresponding authors: matthew.lewis.24@alumni.ucl.ac.uk, s.denaxas@ucl.ac.uk}\textsuperscript{,}\thanks{These authors contributed equally}}
\author[1, 2, 7]{Samuel Thio\textsuperscript{\dag}} 
\author[3]{Amy Roberts}
\author[4]{Catherine Siju}
\author[5]{Whoasif Mukit}
\author[6, 7]{Rebecca Kuruvilla}
\author[2, 6, 7]{Zhangshu Joshua Jiang}
\author[2, 6, 7]{Niko Möller-Grell}
\author[7, 8]{Aditya Borakati}
\author[1, 2, 9]{Richard JB Dobson}
\author[1, 10, 11, 12]{Spiros Denaxas\protect\footnotemark[1]}

\affil[1]{Institute of Health Informatics, University College London, London, U.K.}
\affil[2]{Department of Biostatistics and Health Informatics, King's College London, London, U.K.}
\affil[3]{Imperial College Healthcare NHS Trust, U.K.}
\affil[4]{Black Country Healthcare NHS Foundation Trust, U.K.}
\affil[5]{Feldon Lane Surgery, The Dudley Group NHS Foundation Trust, U.K.}
\affil[6]{The Cleveland Clinic, London, U.K.}
\affil[7]{EPSRC DRIVE-Health CDT, London, U.K.}
\affil[8]{Roger Williams Institute of Liver Studies, School of Immunology and Microbial Sciences, King's College Hospital, London, U.K.}
\affil[9]{CogStack Limited, London, U.K.}
\affil[10]{Interdisciplinary Transformation University (IT:U), Linz, Austria}
\affil[11]{British Heart Foundation Data Science Centre, London, U.K.}
\affil[12]{National and Kapodistrian University of Athens, Athens, Greece}

\date{}

\newcommand{\keywords}[1]{%
  \vspace{1em}
  \noindent \textbf{Keywords:} #1
}

\cftpagenumbersoff{figure}
\cftpagenumbersoff{table}

\begin{document} 
\maketitle
\pagenumbering{roman} 

\vspace{-1cm}

\section*{Abstract:}
\vspace{3pt}

This paper presents the development and evaluation of a Retrieval-Augmented Generation (RAG) system for querying the United Kingdom's National Institute for Health and Care Excellence (NICE) clinical guidelines using Large Language Models (LLMs). The extensive length and volume of these guidelines can impede their utilisation within a time-constrained healthcare system, a challenge this project addresses through the creation of a system capable of providing users with precisely matched information in response to natural language queries. The system's retrieval architecture, composed of a hybrid embedding mechanism, was evaluated against a corpus of 10,195 text chunks derived from three hundred guidelines. It demonstrates high performance, with a Mean Reciprocal Rank (MRR) of 0.814, a Recall of 81\% at the first chunk and of 99.1\% within the top ten retrieved chunks, when evaluated on 7901 queries.\\

The most significant impact of the RAG system was observed during the generation phase. When evaluated on a manually curated dataset of seventy question-answer pairs, RAG-enhanced models showed substantial gains in performance. Faithfulness, the measure of whether an answer is supported by the source text, was increased by 64.7 percentage points to 99.5\% for the RAG-enhanced O4-Mini model and significantly outperformed the medical-focused Meditron3-8B LLM, which scored 43\%. Clinical evaluation by seven Subject Matter Experts (SMEs) further validated these findings, with GPT-4.1 achieving 98.7\% accuracy while reducing unsafe responses by 67\% compared to O4-Mini (from 3.0 to 1.0 per evaluator). This, combined with a perfect Context Precision score of 1 for all RAG-enhanced models, confirms the system's ability to prevent information fabrication by grounding its answers in relevant source material. This study thus establishes RAG as an effective, reliable, and scalable approach for applying generative AI in healthcare, enabling cost-effective access to medical guidelines.

\keywords{Large Language Models, Retrieval-Augmented Generation, NICE Guidelines}

\section{Introduction and Background:}
\vspace{5pt}
This project seeks to develop and evaluate a Retrieval-Augmented Generation (RAG) system \citep{lewis2020retrieval} for querying National Institute for Health and Care Excellence (NICE) clinical guidelines. There are currently thousands of guidelines offered by NICE for clinical, procedural and technological guidance relating to the medical field, with them being viewed millions of times per year \citep{nice_annual}. These guidelines can often be long, with some reaching over a hundred pages. This can lead to a diminished utilisation rate, due to the long time needed to find relevant information in an already overburdened health system \citep{wang_barriers}.\\

The proposed system helps address this problem by enabling users to input a guideline-related query in plain text and receive appropriately matched information, using a Large Language Model (LLM) powered RAG application. We therefore seek to quantify the extent to which RAG enhances the performance of LLMs for querying NICE guidelines.\\

Objectives:
\begin{itemize}
    \item Develop a comprehensive knowledge base by acquiring, pre-processing, and semantically chunking a corpus of NICE guidelines.
    \item Implement and evaluate a variety of retrieval strategies, including sparse, dense, and hybrid search methods, to determine the most effective approach for retrieving relevant excerpts.
    \item Integrate the retrieval system with state-of-the-art LLMs within a RAG architecture, using engineered prompts to ensure responses are contextually grounded.
    \item Conduct a rigorous, two-stage evaluation of the complete RAG system, assessing the performance of both the retrieval and generation components. Benchmark performance against baseline LLMs to quantify improvements in quality and faithfulness.
    \item Evaluate the RAG system outputs with subject matter experts as a measure of real world performance.
\end{itemize}

Through these objectives, this paper will demonstrate a reliable and effective way of integrating advances in generative AI to the healthcare field.\\

\subsection{Natural Language Processing in Healthcare}
\vspace{5pt}
Natural language processing (NLP) has a long history of being applied to the healthcare sector. The ELIZA chatbot \citep{weizenbaum1966eliza}, introduced in 1966 and widely considered the first-ever chatbot, was a rule-based system known for its DOCTOR script simulating a psychotherapist's responses. Furthermore, the PARRY algorithm \citep{colby1971artificial} was introduced in 1971 to mimic the behaviour of a paranoid schizophrenic and is considered one of the first models to partially pass the Turing test \citep{turingcomputing}. While these models offered early insights into the possibilities of NLP in healthcare, their rule-based systems limited their usefulness in real-world settings.\\

\subsection{Large Language Models in Healthcare}
\vspace{5pt}
LLMs have unleashed a new wave of interest in applying chatbots to healthcare problems. Google has been at the forefront of training LLMs specifically for the medical domain with their Med-PaLM \citep{singhal2022large, singhal2025toward}, Med-Gemini \citep{saab2024capabilities} and MedGemma \citep{sellergren2025medgemma} models. These models provide a mix of textual and multimodal capabilities, enabling a wide range of medical query types. They achieve state-of-the-art performance, comparable to that of clinicians, on a variety of tasks.\\

Additionally, open-source initiatives such as the Meditron project, initially started at the EPFL with the release of two fine-tuned Llama-2 models for the medical domain \citep{chen2023meditron}, and now converted to a community-based project fine-tuning open-source models such as the Llama-3 model \citep{sallinen2025llama, dubey2024llama}. They base their fine-tuning on using renowned medical domain sources as training data. This includes a wide variety of clinical guidelines, such as those offered by NICE, along with a variety of PubMed papers. This approach has enabled Meditron models to achieve state-of-the-art performance on benchmarks.\\

While these LLMs show satisfactory results on curated benchmarks, real-world performance is harder to quantify and LLMs have so far lagged in real-world medical implementation. However, a recent study conducted between OpenAI and Penda Health, a private healthcare provider in Kenya, looked at evaluating the effectiveness of a LLM-based clinical decision support tool within a primary care setting \citep{korom2025ai}. This study analysed over thirty-nine thousand patient visits where clinicians were given randomised use of a LLM-powered decision support tool. The system was given context of the patient's consultation purpose as well as summaries of relevant clinical guidelines prior to consultations. Significant reductions in clinical errors were reported, including a decrease of 16\% for diagnostic errors and 12.7\% for treatment errors, highlighting the potential of LLMs to be provided as support tools for clinicians.\\

\subsection{Clinical Guidelines}
\vspace{5pt}
Clinical guidelines are ``systematically developed statements designed to help practitioners and patients make decisions about appropriate health care" \citep{jackson1998guidelines}. These documents serve as one of the cornerstones of evidence-based medicine, providing recommendations that range from concise protocols to comprehensive documents hundreds of pages long. Their primary purpose is to standardise patient care, improve health outcomes and ensure the efficient use of healthcare resources.\\

In the United Kingdom, the National Institute for Health and Care Excellence (NICE) is the principal body responsible for developing these guidelines. Founded in 1999 to ``create consistent guidelines and end rationing of treatment by postcode across the UK" \citep{nice_aboutus_2025}, NICE has become a robust global standard due to its structured and transparent methodology. However, the current manual task of finding information in guidelines can be time-consuming. Recognising this, NICE's five-year plan outlines the need to provide information in ``dynamic, useable formats that support busy health and care practitioners" \citep{nice_five_year} and the need to present recommendations in an ``interactive format". The plan also emphasises the need to rapidly integrate new research and ``make [their] advice and guidance more accessible through the use of technology", a goal this project directly supports.\\

NICE produces diverse types of guidelines. Certain types carry legal implementation requirements; for example, pharmaceutical drugs often require a Technology Appraisal from NICE before being implemented in the NHS. Table \ref{tab:guidance_types} provides a summary of the different guidance types:\\

\begin{longtable}{|p{0.45\linewidth}|p{0.45\linewidth}|}

\hline
\multicolumn{1}{|c|}{\textbf{Current Guidance Types}} &
\multicolumn{1}{c|}{\textbf{Historical Guidance Types}} \\
\hline
\endfirsthead

\hline
\multicolumn{2}{|l|}{\textit{Continued from previous page}} \\
\hline
\multicolumn{1}{|c|}{\textbf{Current Guidance Types}} &
\multicolumn{1}{c|}{\textbf{Historical Guidance Types}} \\
\hline
\endhead

\hline
\multicolumn{2}{|r|}{\textit{Continued on next page}} \\
\endfoot

\hline 
\caption{Current and Historical NICE Guidance Types}
\label{tab:guidance_types} \\
\endlastfoot

\textbf{DG (Diagnostics Guidance):} Guidance on new diagnostic technologies. &
\textbf{CG (Clinical Guideline):} Historical general guidance, now part of NICE guidelines (NG). \\

\textbf{HST (Highly Specialised Technologies Guidance):} Evaluates treatments for very rare conditions. &
\textbf{CSG (Cancer Service Guideline):} Historical guidance created for cancer services. \\

\textbf{IPG (Interventional Procedures Guidance):} Assesses the safety and efficacy of new procedures. &
\textbf{MPG (Medicines Practice Guideline):} Guidance on medical governance. \\

\textbf{MTG (Medical Technologies Guidance):} Guidance on new medical devices and diagnostics. &
\textbf{PH (Public Health Guideline):} Historical guidance on disease prevention, now part of NICE guidelines (NG). \\

\textbf{NG (NICE Guideline):} General guidelines covering clinical, public health and social care. &
\textbf{SC (Social Care Guideline):} Historical guidance for the social care sector, now part of NICE guidelines (NG). \\

\textbf{QS (Quality Standard):} Standards acting as markers for improvement of care. &
\textbf{SG (Safe Staffing Guideline):} Discontinued guidance that made recommendations on NHS staffing levels. \\

\textbf{TA (Technology Appraisal Guidance):} Guidance on the clinical and cost-effectiveness of health technologies such as new medicines, procedures, devices and diagnostic agents. & \\

\end{longtable}

\subsection{Retrieval-Augmented Generation}
\vspace{5pt}
\cite{lewis2020retrieval} introduced RAG to access specific knowledge bases more precisely when using LLMs and reduce the ``hallucination" rates, when a model makes up information in a confident manner. It combines the knowledge LLMs have learned through their pre-training with external knowledge stored in a vector database. The process involves running a similarity search between the user's query and information in the vector database to retrieve relevant content. This retrieved information is then combined with the original query and fed into the LLM's prompt, enabling the model to generate a more informed response.\\

While the majority of LLM implementations in healthcare have focused on question-and-answering tools for examining medical conditions rather than information retrieval, the RAG framework has gained in popularity for medical tasks due to the safety needs of the domain. \cite{macia2024conversational} conduct implementation research on developing a RAG system to query NICE guidelines but do not implement it. This paper thus seeks to build upon the work set out.\\

Implementations of RAG in the medical field include \cite{zakka}, who use a web browser along with an unspecified corpus of medical text and guidelines to pre-retrieve relevant information, which is then passed to a LLM. The system demonstrated an 18\% increase in factuality compared to ChatGPT. Meanwhile, \cite{ferber2024gpt} look at implementing a GPT-4 based RAG system for providing guidance on cancer patient management using oncology guidelines. Findings show that the inclusion of RAG increased the rate of correct responses from 57\% to 84\% with a GPT-4 based RAG implementation. Furthermore, \cite{kresevic2024} showed significant performance improvements from incorporating RAG to interpret Hepatitis C guidelines, with their customized framework increasing accuracy from 43\% to 99\%.\\

A notable implementation for querying UK guidelines is presented by \cite{ive2025}, who developed a clinical Q\&A chatbot for University College London Hospital guidelines using the Llama-3.1-8B model \citep{dubey2024llama}. While the system is not a Retrieval-Augmented Generation (RAG) implementation, it shares similarities, with the system extracting relevant information in response to a query. The key distinction is the system's safety-focused design, which intentionally omits the ``generation" step found in RAG frameworks. To prevent the model from fabricating information, it was uniquely constrained to only extract and display verbatim text from the source guidelines rather than generating new sentences. This method proved highly reliable, achieving 100\% recall for critical information and avoiding ``hallucinated responses". However, the study's scope was limited, focusing only on six small guideline documents, which may have biased its recall performance.\\

In conclusion, while existing research confirms that RAG significantly improves the reliability of guideline related queries passed to LLMs, these applications have been limited to narrow medical specialties or small-scale, non-generative systems. A clear gap remains in evaluating RAG's performance across a broad, national-level corpus. This paper addresses this by implementing and evaluating a RAG system on a wide set of NICE guidelines, aiming to validate its effectiveness for more universal clinical application.\\

\section{Methodology:}
\vspace{5pt}

\subsection{Data}
\vspace{5pt}
The knowledge base for this project was constructed using the complete corpus of guidelines from NICE, acquired via the official NICE API on the 16th of July 2025. In total, 2164 documents were obtained. For this study, the dataset was refined to focus on the most comprehensive guidance types, composed of the 300 NICE Guidelines (NG) and historical Clinical Guidelines (CG) available at the time of acquisition. These documents are extensive, with a mean length of 9611 words, which underscores the challenge of manual information retrieval and highlights the need for an effective automated system.\\

\subsection{Pre-Processing}
\vspace{5pt}
The efficacy of a RAG system is contingent upon the quality and structure of its underlying knowledge base. To this end, a pre-processing pipeline was engineered to transform the raw, heterogeneous NICE guidelines into a refined, queryable resource optimised for retrieval. The native XML format of the data obtained from the API was first converted into Markdown format. This normalisation allows for the preservation of the hierarchical structure of the documents, including headings, lists and tables, while providing a format optimised for ingestion by LLMs.\\

The implementation then used a hierarchical semantic chunking methodology. This approach was selected over naive fixed-size splitting techniques to ensure that the resulting text segments, or ``chunks", were coherent and contextually complete. The strategy operated by first segmenting documents along their major semantic boundaries (i.e. primary and secondary headings).\\

A dual logic was subsequently applied. Oversized chunks exceeding a six hundred token threshold, using the Voyage-3-Large model's tokeniser, were intelligently subdivided at logical breakpoints, such as subsection headings or paragraph breaks, while keeping a fifty-token overlap. Meanwhile undersized chunks below a two hundred token threshold were merged with adjacent segments. This adaptive method of splitting and merging produced passages of optimal length for ingestion by LLMs, mitigating the risks of both context fragmentation and topic dilution.\\

In the final stage, the textual chunks were transformed into retrievable semantic representations through vectorisation. Each chunk was processed by an embedding model, which generated a vector encoding it. These embeddings, along with their corresponding text and metadata, were stored in a database. This created a fully vectorised knowledge base, enabling the RAG system to perform efficient similarity searches across the corpus of clinical guidelines.\\

\subsection{Vector Embeddings}
\vspace{5pt}
We make use of two types of vector embeddings, sparse and dense. The use of these two methods allows for a balance between keyword matching and semantic representation. 

\subsubsection{Sparse Embeddings}
\vspace{5pt}
This project incorporates sparse embeddings to leverage traditional keyword-based retrieval. Sparse embeddings represent documents using high-dimensional vectors where the majority of elements are zero. Each dimension typically corresponds to a specific word in the vocabulary dictionary and the value in that dimension indicates the importance/rarity of the word within the document chunk, with rarer words carrying more importance. This method excels at keyword matching, which is particularly valuable for healthcare, where rare terms are more commonly used. However, it does not consider the semantic meaning of the words being represented or how they relate to nearby terms.\\

For these representations, the Okapi BM25 algorithm \citep{okapibm25}, a widely used and robust ranking function, was employed. BM25 builds upon the Term Frequency-Inverse Document Frequency (TF-IDF) \citep{sparck1972statistical} model to score the relevance of documents to a given query. To improve performance, the text was pre-processed to remove common stop words and to lemmatise the remaining words, ensuring that variations of the same word were treated as a single term. A small Bayesian optimisation was also run for tuning the hyperparameters of the model.\\

The relevance score for a document D given a query Q is calculated as follows:\\

\[\text{score}(D, Q) = \sum_{i=1}^{n} IDF(q_i) \cdot \frac{f(q_i, D) \cdot (k_1 + 1)}{f(q_i, D) + k_1 \cdot \left(1 - b + b \cdot \frac{|D|}{\text{avgdl}}\right)}\]

Where:
\begin{itemize}
  \item \( f(q_i, D) \) is the frequency of term \( q_i \) in document \( D \)
  \item \( |D| \) is the length of document \( D \)
  \item \( \text{avgdl} \) is the average document length in the collection
  \item \( k_1 \) and \( b \) are hyperparameters controlling the term frequency saturation and normalising the scores based on document length
  \item \( IDF(q_i) \) is the inverse document frequency of term \( q_i \)
\end{itemize}

The Inverse Document Frequency (IDF) quantifies the informational value of a word based on its rarity across the document corpus. It is defined as:\\
\[
IDF(q_i) = \log \left( \frac{N - n(q_i) + 0.5}{n(q_i) + 0.5} + 1 \right)
\]

Where:
\begin{itemize}
  \item \( N \) is the total number of documents
  \item \( n(q_i) \) is the number of documents containing term \( q_i \)
\end{itemize}

\subsubsection{Dense Embeddings}
\vspace{5pt}
For this project, a selection of leading dense embedding models was chosen for evaluation. The Voyage-3-Large and Voyage-3.5 models \citep{voyage-3-large2025} were selected for their state-of-the-art performance and their capability to handle long context windows, as shown in Table \ref{tab:model_specs}. The Qwen3-Embedding-0.6B model \citep{zhang2025qwen3} was included due to its place as a top-performing open-source model on the Massive Multilingual Text Embedding Benchmark (MMTEB) leaderboard \citep{enevoldsen2025mmtebmassivemultilingualtext}, a key benchmark for embedding model performance. Finally, OpenAI's text-embedding-3-large model \citep{openaiemb} was chosen due to its widespread popularity.\\

In contrast to the keyword-based approach of sparse embeddings, dense embeddings capture deeper semantic meaning in texts. These embeddings are high-dimensional vector representations generated using transformer-based neural networks, such as the BERT model \citep{devlin2019bert}. By mapping texts into a continuous vector space of fixed length, as shown in Table \ref{tab:model_specs}, these models can represent complex semantic relationships. They are capable of processing long context windows, which is crucial for fully representing the information in chunks.\\ 

The generation of dense embeddings begins with tokenisation, a process where the input text is broken down into smaller units called ``tokens", which can be thought of as numerical representations of words. This segmentation is the crucial first step in converting human-readable language into a numerical format that the model can process. This project uses the tokeniser associated with the state-of-the-art Voyage-3-Large embedding model by Voyage AI.\\

Unlike the pre-processing conducted for the sparse embeddings, stop word removal and lemmatisation were intentionally omitted for the generation of dense embeddings. This omission is critical for preserving the full semantic context of the texts, as transformer models leverage subtle grammatical structures to create more nuanced vector representations.\\

\begin{figure}[h]
\centering
\includegraphics[height=7.5cm]{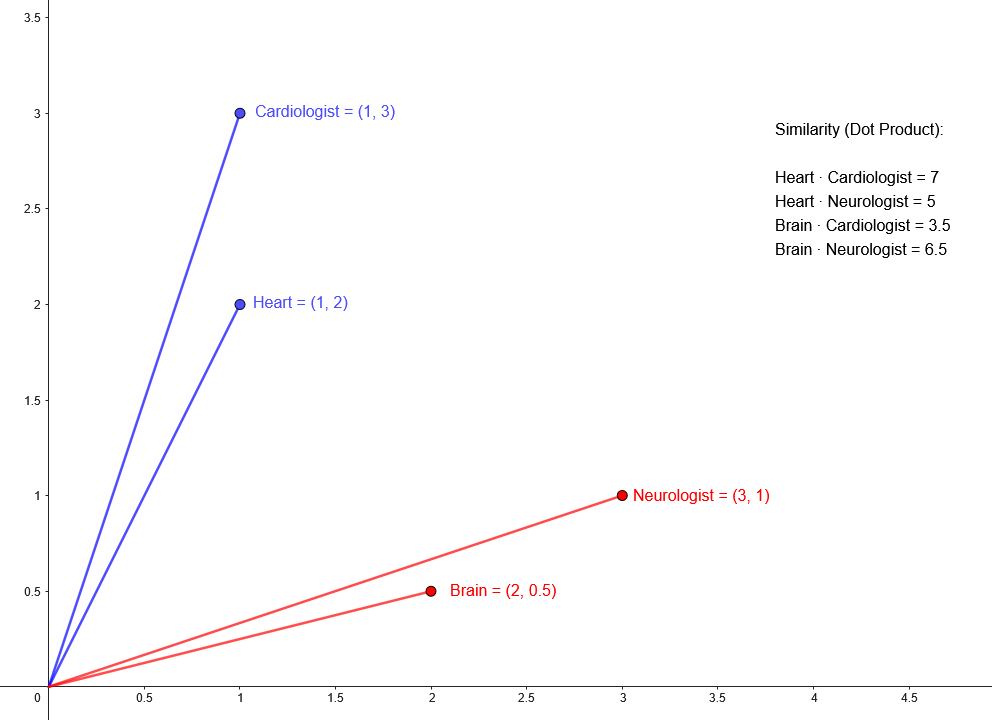}
\caption{Example Vector Representation of Words in 2D}
\label{fig:vector_representation}
\end{figure}

Concepts with similar meanings are thus positioned closer together in vector space. For example, the vectors for ``Heart" and ``Cardiologist" would have a higher similarity score than the vectors for ``Heart" and ``Neurologist", as illustrated in Figure \ref{fig:vector_representation}. This representation enables the system to perform deeper similarity searches that go beyond keyword matching.\\

\begin{table}[h!]
    \centering
    \small
    \renewcommand{\arraystretch}{1.2}
    \resizebox{\textwidth}{!}{
        \begin{tabular}{|l|c|c|c|c|}
            \hline
            \textbf{Model} & \textbf{Provider} & \textbf{Embedding Dimension} & \textbf{Context Length (Tokens)} & \textbf{Availability} \\
            \hline
            Voyage-3-Large & Voyage AI & 2048 & 32 000 & Closed-Source \\
            \hline
            Voyage-3.5 & Voyage AI & 2048 & 32 000 & Closed-Source \\
            \hline
            text-embedding-3-large & OpenAI & 3072 & 8191 & Closed-Source \\
            \hline
            Qwen3-Embedding-0.6B & Alibaba & 1024 & 32 768 & Open-Source \\
            \hline
        \end{tabular}
    }
    \caption{Comparison of Embedding Model Specifications}
    \label{tab:model_specs}
\end{table}

\subsection{Retrieval}
\vspace{5pt}
\subsubsection{Retrieval Method}
\vspace{5pt}
The core of the dense embedding retrieval process is a similarity search performed on the high-dimensional vector space. When a user submits a query, it is first converted into a dense vector using the same embedding model that processed the original guideline chunks. This ensures that both the query and relevant excerpts are represented in the same semantic space.\\

The system then calculates the similarity between the user's query vector and every guideline chunk vector stored in the database. To ensure that this comparison measures semantic similarity rather than being influenced by vector magnitude, all vectors are pre-normalised to have a unit length of one during the embedding process. As a result, the computationally efficient dot product can be used for retrieval. A higher score indicates a greater degree of semantic similarity between the query and the guideline chunk.\\

\vspace{0.3cm}
The dot product is defined as:
\[\text{Dot Product} = \mathbf{A} \cdot \mathbf{B} = \sum_{i=1}^n a_i \cdot b_i\]
Where:
\begin{itemize}
    \item \(\mathbf{A}\) represents the query vector.
    \item \(\mathbf{B}\) represents a guideline chunk vector.
    \item \(a_i\) is the \(i\)-th element of vector \(\mathbf{A}\).
    \item \(b_i\) is the \(i\)-th element of vector \(\mathbf{B}\).
    \item \(n\) is the total number of dimensions in the vectors.
\end{itemize}
\vspace{0.5cm}
Once the dot product is calculated for all pairs, the system ranks them in descending order based on their similarity scores, with the highest scoring matches indicating topic relevance and being passed on to the completion model. Table \ref{tab:example_dot} provides an illustration of this principle: a query about ``catheter-directed thrombolytic therapy" would be expected to achieve a high similarity score with a relevant guideline section, while its similarity to an irrelevant context on Hormone Replacement Therapy would be negligible.\\

\begin{table}[h!]
\small
\centering
\renewcommand{\arraystretch}{1.5} 
\begin{tabular}{|p{0.35\textwidth}|p{0.48\textwidth}|p{0.1\textwidth}|}
\hline
\textbf{Question} & \textbf{Context} & \textbf{Similarity} \\
\hline
Should catheter-directed thrombolytic therapy be considered for people with symptomatic iliofemoral deep vein thrombosis and symptoms for the past month? 
& 
NG158 Section 1.6.1 Consider catheter-directed thrombolytic therapy for people with symptomatic iliofemoral DVT who have: symptoms lasting less than 14 days, good functional status, a life expectancy of 1 year or more, and a low risk of bleeding.
& 
0.95 \\
\hline
Should catheter-directed thrombolytic therapy be considered for people with symptomatic iliofemoral deep vein thrombosis and symptoms for the past month? 
& 
NG23 Section 1.6.1: When discussing HRT as a treatment option for menopause-associated symptoms, explain that, overall, taking either combined HRT or oestrogen-only HRT is unlikely to affect life expectancy.
& 
0.15 \\
\hline
\end{tabular}
\caption{Example of Good vs. Poor Context Matching}
\label{tab:example_dot}
\end{table}

\newpage
\subsubsection{Hybrid Search}
\vspace{5pt}
To leverage the complementary strengths of both sparse and dense retrieval, a hybrid search approach is implemented using a Weighted Reciprocal Rank Fusion \citep{cormack2009reciprocal}. This technique normalises ranks across different retrievals into a single score, so that passages ranked highly across multiple models are assigned higher relevance.\\

The final weighted score for each document is calculated as follows:\\

\[WRRF_{doc} = \sum_m (\frac{w_m}{k + rank_{m,doc}})\]
\begin{itemize}
    \item \(WRRF_{doc}\) is the final weighted reciprocal rank score for the document.
    \item \(m\) represents a specific model in the ensemble.
    \item \(w_m\) is the weight assigned to the model \(m\).
    \item \(rank_{m,doc}\) is the rank of the document using model \(m\).
    \item \(k\) is a hyperparameter which controls the influence level of documents with higher ranks.
\end{itemize}
\vspace{0.5cm}
By combining the keyword-matching precision of sparse embeddings with the contextual understanding of dense embeddings, this hybrid approach aims to improve overall retrieval accuracy, addressing some of the limitations of each method.\\

\subsubsection{Reranking}
\vspace{5pt}
To enhance the precision of the retrieved results, a reranking stage is incorporated into the retrieval pipeline. This process uses a cross-encoder architecture \citep{nogueira2019passage}, which differs from the bi-encoder model employed in the initial retrieval phase. Rerankers have been shown to lead to higher precision than bi-encoders \citep{rosa2022defense}; however, they come at a higher computational cost. Their use is therefore restricted to the subset of top initially retrieved chunks to improve performance.\\

\begin{figure}[ht]
\centering
\includegraphics[height=10cm]{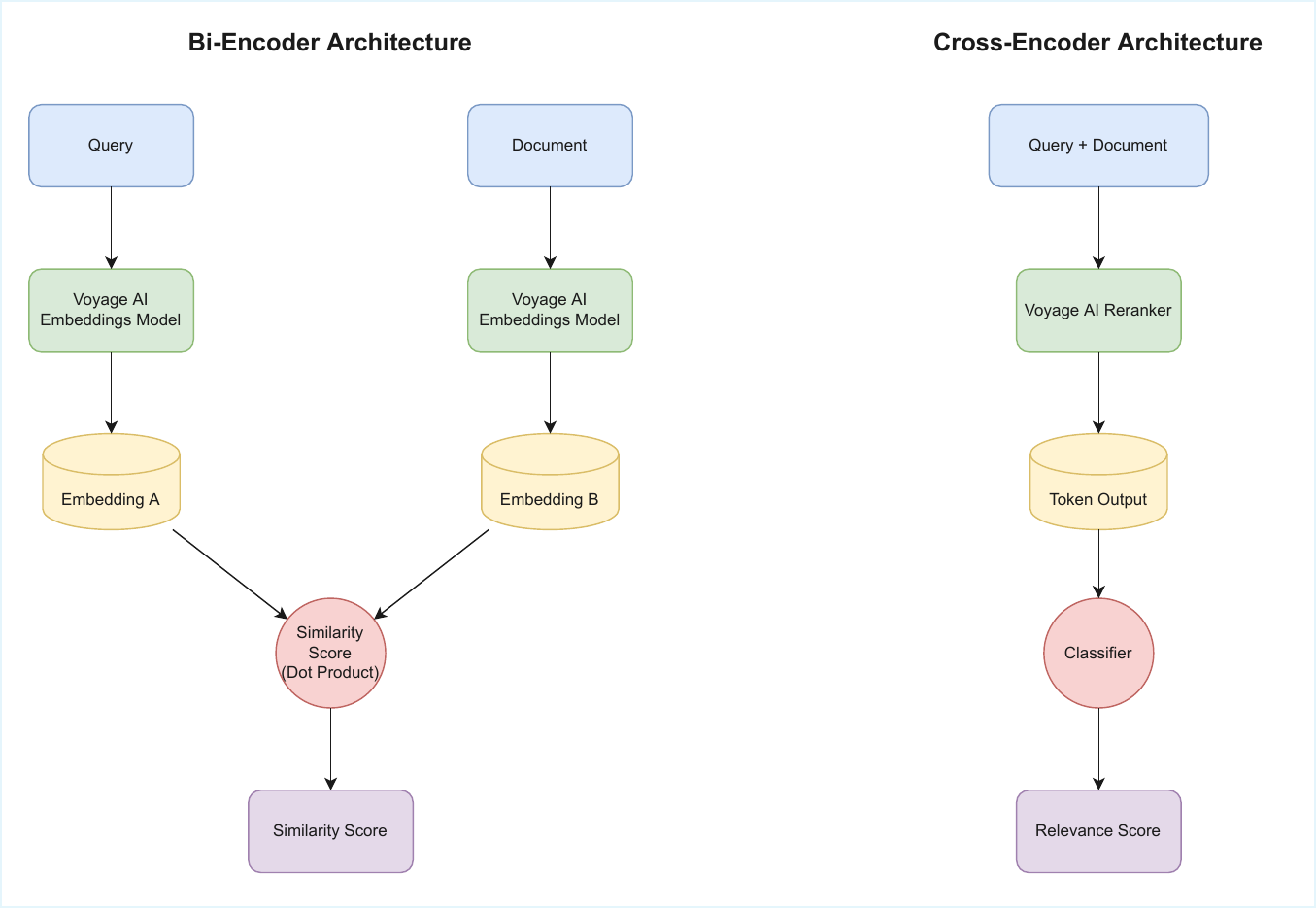}
\caption{Bi-Encoder Vs. Cross-Encoder Architecture}
\label{fig:architecture_diagram}
\end{figure}

As illustrated in Figure \ref{fig:architecture_diagram}, the bi-encoder architecture independently generates vector representations for the query and document. In contrast, the cross-encoder architecture jointly encodes the query and a candidate document by concatenating them into a single input sequence. This combined representation is then processed by the model, which subsequently assigns a relevance score indicating the likelihood that the document accurately addresses the query. This method allows for a more nuanced and context-aware assessment of relevance.\\

\vspace{1cm}

\subsection{LLM Completions}
\vspace{5pt}
The completion phase of the RAG system relies on LLMs and a carefully designed prompt engineering framework. This section details the models and prompts used to generate accurate, context-aware, and correctly formatted responses.\\

\subsubsection{Model Selection and Configuration}
\vspace{5pt}
Several advanced LLMs were selected for evaluation, including OpenAI’s GPT-4.1 \citep{openai_gpt41_2024} and O4-Mini \citep{openai_o3_o4mini_systemcard_2024} models, alongside Anthropic's Claude Sonnet 4 \citep{anthropic2025claude4}. These models are chosen for their different strengths, from the efficiency of GPT-4.1-nano to the advanced reasoning capabilities of O4-Mini. A key advantage of these models is their ability to process long context windows, with the GPT-4.1 family supporting up to one million tokens and both O4-Mini and Claude Sonnet 4 supporting two hundred thousand tokens, as shown in Table \ref{tab:llm_models}. This is critical for ingesting the large amount of contextual information passed from the database's context chunks.\\

To make the system's outputs as deterministic and reproducible as possible, the temperature parameter of the models is set to 0 where possible. This setting minimises randomness in responses, which helps constrain model outputs to be strictly based on the provided context and thus reduce the chance of hallucinations.\\
\vspace{7cm}
\begin{table}[h!]
    \centering
    \small
    \renewcommand{\arraystretch}{1.2}  
    \resizebox{\textwidth}{!}{
        \begin{tabular}{|l|l|c|c|c|}
            \hline
            \textbf{Model} & \textbf{Provider} & \textbf{Year of Release} & \textbf{Context Window (Tokens)} & \textbf{Reasoning} \\
            \hline
            GPT-4.1 & OpenAI & 2025 & 1 000 000 & No \\
            \hline
            GPT-4.1-Mini & OpenAI & 2025 & 1 000 000 & No \\
            \hline
            GPT-4.1-Nano & OpenAI & 2025 & 1 000 000 & No \\
            \hline
            O4-Mini & OpenAI & 2025 & 200 000 & Yes \\
            \hline
            Claude Sonnet 4 & Anthropic & 2025 & 200 000 & Yes \\
            \hline
            Meditron3-8B & Meditron Project & 2024 & 8192 & No \\
            \hline
        \end{tabular}
    }
    \caption{Comparison of Large Language Models Used for Generation}
    \label{tab:llm_models}
\end{table}

\subsubsection{Prompt Engineering}
\vspace{5pt}

The generation phase of the LLMs is guided by a two-part prompt structure, consisting of a ``System Prompt" and a ``User Prompt", as shown in Table \ref{tab:non_rag_prompt} and Table \ref{tab:system_prompt}.\\

\textbf{System Prompt}:
\vspace{0.2cm}
\begin{itemize}
    \item \textbf{Contextual Restriction:} The model is prohibited from generating information outside of the supplied context for RAG models. This ensures faithfulness and prevents the fabrication of information. Meanwhile non-RAG models are instructed to only use information from NICE guidelines.

    \item \textbf{Formatting and Informational Fidelity:} The prompt gives explicit formatting instructions to ensure readable and well-structured output. This includes rules for using markdown for lists and for preserving any tables from the source context, as well as preserving URLs.

    \item \textbf{Handling of Null Context:} As a safeguard, the model is instructed to give a specific response of ``No relevant NICE guidelines were found" if the retrieved context contains no relevant information to answer the query. This prevents the model from trying to formulate an answer when no relevant evidence is retrieved.
\end{itemize}
\vspace{10pt}
\textbf{User Prompt}:\\

The user prompt is the dynamic input for the system, which changes for each request. It uses a placeholder \{query\_text\} which is populated with the user's question, while for RAG models a placeholder \{context\_text\} is also populated with relevant text chunks retrieved from the retrieval pipeline.\\

\begin{table}[H]
    \footnotesize 
    \centering
    \renewcommand{\arraystretch}{1.2} 
    \begin{tabularx}{\textwidth}{lX}
        \toprule
        \textbf{Role} & \textbf{Content / Instructions} \\
        \midrule
        System &
        You are a medical AI assistant tasked with answering clinical questions strictly based on NICE clinical guidelines. Follow these rules:
        
        \begin{enumerate}[label=\arabic*., wide, labelindent=0pt, itemsep=1pt, topsep=2pt]
            \item Only use information from NICE guidelines.
            \item If no relevant NICE guideline information is available, reply: `No relevant NICE guidelines were found.'
            \item Be concise. Use markdown for lists and tables.
            \item Never fabricate sources or references.
        \end{enumerate} \\
        \midrule
        User &
        \texttt{\{query\_text\}} \\
        \bottomrule
    \end{tabularx}
    \caption{System and User Prompt for Non-RAG LLMs}
    \label{tab:non_rag_prompt}
\end{table}
\vspace{5pt}

\begin{table}[H]
    \footnotesize 
    \centering
    \renewcommand{\arraystretch}{1.2} 
    \begin{tabularx}{\textwidth}{lX}
        \toprule
        \textbf{Role} & \textbf{Content / Instructions} \\
        \midrule
        
        System &
        You are a medical AI assistant tasked with answering clinical questions strictly based on the provided NICE clinical guidelines context. Follow the requirements below to ensure accurate, consistent, and professional responses.
        
        \textbf{\# Response Rules}
        
        \begin{enumerate}[label=\arabic*., wide, labelindent=0pt, itemsep=1pt, topsep=2pt]
            \item \textbf{Context Restriction}:
            \begin{itemize}[label=\textendash, itemsep=0pt, topsep=1pt, leftmargin=*]
                \item Only use information given in the provided NICE guidelines context.
                \item Do not generate or speculate with information not explicitly found in the given context.
            \end{itemize}

            \item \textbf{Answer Format}:
            \begin{itemize}[label=\textendash, itemsep=0pt, topsep=1pt, leftmargin=*]
                \item Provide a clear and concise response based solely on the context.
                \item When including a list, use standard markdown bullet points (`*` or `-`).
                \item If a list follows introductory text, insert a line break before the first bullet point.
                \item Each bullet point must be on its own line.
            \end{itemize}

            \item \textbf{Preserve Tables}:
            \begin{itemize}[label=\textendash, itemsep=0pt, topsep=1pt, leftmargin=*]
                \item If relevant markdown tables appear in the context, reproduce them in your answer.
                \item Maintain the original structure, formatting, and content of any included tables.
            \end{itemize}

            \item \textbf{Links and URLs}:
            \begin{itemize}[label=\textendash, itemsep=0pt, topsep=1pt, leftmargin=*]
                \item Include any URLs or web links from the context directly in your response when relevant.
                \item Integrate links naturally within sentences, using markdown syntax for clickable text links.
                \item DO NOT generate or invent any URLs not explicitly present in the context.
            \end{itemize}
            
            \item \textbf{Markdown Link Formatting}:
            \begin{itemize}[label=\textendash, itemsep=0pt, topsep=1pt, leftmargin=*]
                \item In responses, only the descriptive text in brackets should be visible and clickable (e.g., `[NICE...](https://...)`).
                \item Readers should never see raw URLs in the text.
            \end{itemize}

            \item \textbf{If No Relevant Information}:
            \begin{itemize}[label=\textendash, itemsep=0pt, topsep=1pt, leftmargin=*]
                \item If the context contains no relevant information, state clearly: \newline
                \textit{``No relevant NICE guidelines were found."}
            \end{itemize}
        \end{enumerate}

        \textbf{\# Output Format}
        \begin{itemize}[label=\textendash, itemsep=0pt, topsep=1pt, leftmargin=*]
            \item All responses should be in plain text, using markdown formatting for lists and links.
            \item Do not use code blocks.
            \item Answers should be concise, accurate, and formatted according to the rules above.
        \end{itemize}
        
        \textbf{\# Examples}
        \begin{itemize}[label={}, itemsep=1pt, topsep=1pt, leftmargin=*]
            \item \textbf{Example 1: Integration of markdown link in context} \newline
            \texttt{Question: "What is the recommended treatment for stage 2 hypertension?"} \newline
            \texttt{Context snippet: ...see the [NICE hypertension guidelines](https://...)} \newline
            \texttt{Output:} \newline 
            \texttt{According to the [NICE hypertension guidelines](https://...), stage 2...}

            \item \textbf{Example 2: Multiple guideline references} \newline
            \texttt{According to these guidelines:} \newline
            \texttt{* Initial treatment is lifestyle modification.} \newline
            \texttt{* For persistent hypertension, refer to [hypertension...](https://...)}.

            \item \textbf{Example 3: No relevant context} \newline
            \texttt{No relevant NICE guidelines were found.}
        \end{itemize}
        
        \textbf{\# Notes}
        \begin{itemize}[label=\textendash, itemsep=0pt, topsep=1pt, leftmargin=*]
            \item Never output information beyond what is provided in the supplied context.
            \item Always use markdown for lists and links.
            \item Ensure all relevant markdown tables are preserved in your answer.
            \item Present links only as clickable text, not as bare URLs.
        \end{itemize}
        
        \textbf{REMINDER:} Strictly adhere to all formatting and content rules above.
        \\ \midrule

        User &
        \texttt{\{query\_text\}}
        \newline\newline
        \texttt{Context from NICE clinical guidelines:}
        \newline
        \texttt{\{context\_text\}}
        \\ \bottomrule
    \end{tabularx}
    \caption{System and User Prompt for RAG LLMs}
    \label{tab:system_prompt}
\end{table}

\subsection{Testing}
\vspace{5pt}
To rigorously evaluate the efficacy of the developed RAG system, a comprehensive, two-stage testing methodology was implemented. This approach facilitates the isolated assessment of each core component, retrieval and generation, thereby allowing for a granular analysis of their respective performances. The first stage focuses on the retrieval system's ability to source relevant information, while the second stage evaluates the quality of the final generated answers using the specialised Retrieval Augmented Generation Assessment (RAGAs) framework \citep{ragas}.\\

\subsubsection{Retrieval Component Evaluation}
\vspace{5pt}
The primary objective of this stage is to measure the effectiveness of the document retrieval system in identifying the most relevant text chunks from the knowledge base in response to a user query. A synthetic dataset of queries was generated to form a ground truth for evaluation. The process began with extracting 10,195 chunks from the 300 NICE guidelines used.\\

These chunks were then filtered to remove boilerplate sections such as ``Update information" or ``Committee members", ensuring that only clinically relevant content was retained for generating queries. GPT-4.1-Nano was employed to generate realistic, high-quality queries that a healthcare professional might use to retrieve information. This was achieved through a carefully engineered prompt, as detailed in Table \ref{tab:llm_prompts}, which instructed the model to create a question directly addressing each provided text chunk. This automated process resulted in a comprehensive evaluation dataset of 9296 query/chunk pairs, where chunks were each paired with a corresponding synthetic query. This dataset was then partitioned into a validation set (15\%), used for BM25 hyperparameter tuning, and a testing set (85\%) for the final evaluation.\\

\begin{table}[H]
    \small
    \centering
    \renewcommand{\arraystretch}{1.3} 
    \begin{tabularx}{\textwidth}{lX}
        \toprule
        \textbf{Prompt Type} & \textbf{Content} \\
        \midrule
        \textbf{System Prompt} & 
        You are an expert assistant specialized in generating realistic search queries for NICE guidelines.
        
        \textbf{Your task:} Generate a natural question that a healthcare professional or patient would realistically use to find the given information.
        
        \textbf{Requirements:}
        \begin{itemize} \itemsep0pt 
            \item Start with question words: `What', `How', `When', `Should', etc.
            \item Focus on the core clinical topic or specific recommendation.
            \item Use natural medical terminology that real users would search for.
            \item Keep it concise and directly related to the content.
            \item Return only the question, no additional formatting or explanations.
        \end{itemize}
        
        \textbf{Examples:}
        \begin{itemize} \itemsep0pt
            \item ``What are the treatment options for managing hypertension in pregnant women?"
            \item ``How should blood glucose be monitored in diabetes patients?"
            \item ``When should antibiotics be prescribed for respiratory infections?"
        \end{itemize} \\
        \midrule
        \textbf{User Prompt} & 
        Document Excerpt: \texttt{\{doc\_content\}}
        
        Generate a realistic search query for this NICE guideline content.\\
        \bottomrule
    \end{tabularx}
    \caption{Testing Question Generation Prompts}
    \label{tab:llm_prompts}
\end{table}
\newpage
\underline{\textbf{Retrieval Evaluation Metrics}}: the performance of each retrieval configuration is quantified using the following metrics:

\begin{itemize}
\item \textbf{MRR} (Mean Reciprocal Rank): The average of the reciprocal ranks for a set of queries, measuring how high correct chunks are ranked. A higher value indicates the correct chunk is found closer to the top.
\item \textbf{Recall@k}: The proportion of queries where the correct chunk is retrieved in the top k results. A higher value indicates better performance at retrieving relevant items within the top k positions.
\item \textbf{Median Rank}: The median position of the correct chunk across all queries. A lower value indicates that for a majority of queries, the correct chunk is found at or below the median rank.
\item \textbf{Mean Rank}: The average position of the correct chunk across all queries. A lower value signifies that, on average, relevant items are ranked higher.
\item \textbf{Max Rank}: The highest rank of a correct chunk observed across all queries. A lower value indicates better performance for the worst retrieval.
\end{itemize}

\subsubsection{LLM Completion Evaluation}
\vspace{5pt}
The \mbox{RAGAs} framework, an evaluation tool that leverages LLMs to assess the performance of RAG pipelines, was used to evaluate the quality and reliability of the generated answers in the completion phase. It evaluates the LLM's ability to extract relevant information from the retrieved context along with its faithfulness. The evaluation was performed using a manually curated dataset of seventy question-answer pairs derived from the NICE guidelines, an example of which is shown in Table \ref{tab:sample_qa}.\\

\begin{table}[!ht]
    \small
    \centering
    \renewcommand{\arraystretch}{1.3} 
    \begin{tabular}{|p{4cm}|p{10cm}|p{2cm}|}
        \hline
        \textbf{Question} & \textbf{Answer} & \textbf{Matched Source} \\
        \hline
        What factors identify babies as being at increased risk of developing significant hyperbilirubinaemia? & 
        Identify babies as being more likely to develop significant hyperbilirubinaemia if
        they have any of the following factors:
        \begin{itemize}
            \item gestational age under 38 weeks
            \item a previous sibling with neonatal jaundice requiring phototherapy
            \item mother's intention to breastfeed exclusively
            \item visible jaundice in the first 24 hours of life. 
        \end{itemize} & 
        \makecell[l]{CG98, \\ Section 1.2} \\
        \hline
    \end{tabular}
    \caption{Sample QA Pair from Testing Dataset}
    \label{tab:sample_qa}
\end{table}

The framework uses a LLM, in this case GPT-4.1-Mini, to judge the performance of the system, assessing the generated responses against ground-truth answers and the retrieved context. The source code of the framework was modified to make system prompts better suited to the medical domain. The following four metrics from the RAGAs package were employed to provide a multi-faceted view of the system's performance:\\

\begin{itemize}
    \item \textbf{Context Precision with Reference}: Measures the proportion of chunks in the retrieved context which are relevant to the user query.
    \item \textbf{Context Recall}: Measures the proportion of relevant chunks successfully retrieved.
    \item \textbf{Response Relevancy}: Verifies the answer addresses the query in an appropriate way.
    \item \textbf{Faithfulness}: Verifies that the claims made in the answer can be inferred from the context.
\end{itemize}

To quantify the value added by the retrieval architecture, the RAG-enabled system was benchmarked against a variety of baseline LLMs operating without access to the curated vector database. These baseline models included Claude Sonnet 4, GPT-4.1, GPT-4.1-Mini, GPT-4.1-Nano, O4-Mini and the medically focused Meditron3-8B.\\

The non-RAG models were evaluated under two different scenarios to measure the RAG system's impact on accuracy and reliability. Firstly, the models were evaluated using only their pre-trained knowledge. Secondly, a more powerful baseline was created by testing Claude Sonnet 4 with a web search function limited to the nice.org.uk domain, the same information source used by the RAG system.\\

\subsection{Clinical Evaluation by Subject Matter Experts}
\vspace{5pt}
In addition to the automated evaluation metrics, a clinical evaluation of the NICE-RAG system was conducted by a panel of Subject Matter Experts (SMEs), consisting of 7 NHS clinicians from various medical specialities (General Practice, Emergency Medicine, General Surgery, Psychiatry, General Medicine and Neurology). The evaluators were tasked with assessing the system's performance on a dataset derived from the NICE clinical guidelines, consisting of the same seventy question-answer pairs evaluated used for RAGAs evaluation.\\

The evaluation criteria were twofold:
\begin{enumerate}
    \item \textbf{Accuracy Score}: Evaluators scored the system's outputs on a scale:
    \begin{itemize}
        \item \textbf{0}: Wrong answer which did not answer the query and had wrong context
        \item \textbf{0.5}: Partial answer which answered the query partially and had incomplete context
        \item \textbf{1}: Complete answer which answered the query and contained full context
    \end{itemize}
    
    \item \textbf{Safety Score}: A binary assessment of the output's safety:
    \begin{itemize}
        \item \textbf{0}: Safe
        \item \textbf{1}: Unsafe
    \end{itemize}
\end{enumerate}

The total number of unsafe responses was summed for each evaluator to provide an aggregate safety metric. The evaluation compared the performance of two Large Language Models: O4-Mini and GPT-4.1.\\

\section{Results:}
\vspace{5pt}
This section presents the results of the two-stage evaluation process, focusing firstly on the performance of the retrieval system and secondly on the quality and reliability of the answers generated from the complete RAG pipeline.\\

\subsection{Retrieval Results}
\vspace{5pt}
The retrieval system's effectiveness was measured by its ability to identify the correct document chunk from a database of 10,195 chunks in response to 7901 synthetically generated queries. The performance of various embedding models and retrieval strategies are detailed in Table \ref{tab:retrieval_metrics}.\\
\newpage
\begin{table}[h!]
    \centering
    \small
    \renewcommand{\arraystretch}{1.2}
    \resizebox{\textwidth}{!}{
        \begin{tabular}{|l|c|c|c|c|c|c|c|c|}
            \hline
            \textbf{Model} & \textbf{MRR} & \textbf{Recall@1} & \textbf{Recall@5} & \textbf{Recall@10} & \textbf{Recall@15} & \textbf{Median Rank} & \textbf{Mean Rank} & \textbf{Max Rank} \\
            \hline
            Voyage-3-Large & \textbf{0.826} & 0.718 & 0.962 & 0.985 & 0.993 & 1 & 1.836 & 251 \\
            \hline
            Voyage-3.5 & 0.788 & 0.665 & 0.943 & 0.978 & 0.987 & 1 & 2.236 & 262 \\
            \hline
            Text-Embedding-3-Large & 0.749 & 0.615 & 0.924 & 0.970 & 0.983 & 1 & 2.571 & 292 \\
            \hline
            Qwen3-Embedding-0.6B & 0.776 & 0.653 & 0.933 & 0.973 & 0.984 & 1 & 2.697 & 767  \\
            \hline
            BM25 & 0.625 & 0.482 & 0.806 & 0.887 & 0.924 & 2 & 14.151 & 9908 \\
            \hline
            Voyage-3-Large + BM25 & 0.814 & 0.699 & 0.960 & 0.989 & \textbf{0.995} & 1 & 1.829 & 185 \\
            \hline
            Voyage-3-Large + Text-Embedding-3-Large & 0.819 & 0.707 & 0.960 & 0.988 & 0.994 & 1 & \textbf{1.810} & \textbf{70} \\
            \hline
            \multicolumn{9}{|c|}{\textit{With Voyage Reranker-2-Lite}} \\
            \hline
            Voyage-3-Large + BM25 & - \footnotemark[1]& 0.779 & 0.977 & 0.990 & \textbf{0.995} & - \footnotemark[1]& - \footnotemark[1]& - \footnotemark[1]\\
            \hline
            \multicolumn{9}{|c|}{\textit{With Voyage Reranker-2}} \\
            \hline
            Voyage-3-Large + BM25 & - \footnotemark[1]& \textbf{0.810} & \textbf{0.982} & \textbf{0.991} & \textbf{0.995} & - \footnotemark[1]& - \footnotemark[1]& - \footnotemark[1]\\
            \hline
        \end{tabular}
    }
    \caption{Retrieval Results from 7901 Queries Run Against a Database of 10195 Chunks}
    \label{tab:retrieval_metrics}
\end{table}

\footnotetext[1]{Metrics omitted due to prohibitive API costs. The reranker's $O(n \cdot k)$ API call complexity, with n the number of queries (7901) and k the number of query/chunk pairs evaluated (10195), was too expensive to run. Only the top 15 chunks are therefore reranked for each query, as is standard for reranker usage. Observed recall gains imply all rank-based metrics would also improve.}

The results show that dense embedding models significantly outperform the traditional sparse embedding BM25 model. The top-performing individual model was Voyage-3-Large, achieving a MRR of 0.826, mean rank of 1.836, and retrieving the correct chunk within the top fifteen results for 99.3\% of queries. The model's effectiveness is highlighted by its ability to retrieve the correct document, from a corpus of 10,195 chunks, as the first ranked result for 71.8\% of queries (Recall@1). In contrast, the BM25 baseline was far less effective, with an MRR of 0.625 and a much higher mean rank of 14.151.\\

Hybrid search methods, which combine the results of multiple models, demonstrated improved performance in certain areas. Combining the Voyage-3-Large and Text-Embedding-3-Large models yielded the lowest maximum rank of all tested methods. Meanwhile, a hybrid of Voyage-3-Large with BM25, combined with the Voyage Reranker-2, achieved the highest overall Recall@1 of 81\% and Recall@10 of 99.1\%, indicating its strength in consistently placing relevant documents high in the rankings, as well as validating rerankers as a tool for improving precision.\\

In contrast, the open-source Qwen3-Embedding-0.6B model performed less effectively than most closed-source counterparts. Its max rank was significantly higher than other dense models. However, it outperformed the OpenAI's text-embedding-3-model across all other metrics, indicating that open-source models can achieve competitive performance.\\

\subsection{Completion Results}
\vspace{5pt}

The completion evaluation assessed the final output of the RAG system against baseline non-RAG models using a curated set of seventy question-answer pairs. The top ten reranked results, using the Voyage-3-Large + BM25 retrieval combination, were passed as context to the models. The results, shown in Table \ref{tab:ragas_results}, highlight the critical role of the RAG architecture in ensuring the safety and reliability of the generated answers.\\

All RAG-enabled models achieved perfect Context Precision and Context Recall scores of 1.0 when passing the top ten retrieved chunks as context, with slightly lower Context Recall observed when passing the top five chunks, thus confirming that the retrieval stage effectively sources all relevant information. This directly translates to strong performance in Faithfulness, with the RAG-enabled O4-Mini achieving a near-perfect score of 0.995, an increase of 0.647 from its non-RAG counterpart.\\

This stands in stark contrast to the baseline non-RAG models, which exhibited a high propensity for ``hallucinations". The medically-focused Meditron3-8B model scored just 0.430 on the Faithfulness score, while the GPT-4.1 and Claude Sonnet 4 models, which are considered two of the leading LLMs, scored 0.596 and 0.589 respectively. Even when augmented with domain specific web-search capabilities, the Claude Sonnet 4 model's performance, while improved, still fell short of the RAG system's reliability. This demonstrates that without the grounding context provided by RAG pipelines, even state-of-the-art LLMs are unreliable for grounded clinical query answering.\\

\begin{table}[h!]
    \centering
    \small 
    \renewcommand{\arraystretch}{1.2}
    \begin{tabular}{|p{4cm}|c|c|c|c|}
        \hline
        \textbf{Model} & \makecell{\textbf{Context} \\ \textbf{Precision}} & \makecell{\textbf{Context} \\ \textbf{Recall}} & \makecell{\textbf{Response} \\ \textbf{Relevancy}} & \textbf{Faithfulness} \\
        \hline
        \multicolumn{5}{|c|}{\textit{Baseline Non-RAG LLM}} \\
        \hline
        Claude Sonnet 4 & - \footnotemark[2] & - \footnotemark[2] & 0.805 & 0.589 \\
        \hline
        Claude Sonnet 4 with Web-Search & - \footnotemark[2] & - \footnotemark[2]& 0.781 & 0.883 \\
        \hline
        GPT-4.1 & - \footnotemark[2] & - \footnotemark[2] & 0.750 & 0.596 \\
        \hline
        GPT-4.1-Mini & - \footnotemark[2] & - \footnotemark[2] & 0.793 & 0.566 \\
        \hline
        GPT-4.1-Nano & - \footnotemark[2] & - \footnotemark[2] & 0.453 & 0.550 \\
        \hline
        O4-Mini & - \footnotemark[2] & - \footnotemark[2] & 0.704 & 0.348 \\
        \hline
        Meditron3-8B & - \footnotemark[2] & - \footnotemark[2] & 0.851 & 0.430 \\
        \hline        
        \multicolumn{5}{|c|}{\textit{RAG}} \\
        \hline
        Claude Sonnet 4 @ 10 & 1 & 1 & 0.868 & 0.991 \\
        \hline
        Claude Sonnet 4 @ 5 & 1 & 0.992 & 0.863 & 0.989 \\
        \hline
        GPT-4.1 @ 10 & 1 & 1 & 0.858 & 0.993 \\
        \hline
        GPT-4.1-Mini @ 10 & 1 & 1 & \textbf{0.878} & 0.985 \\
        \hline
        GPT-4.1-Nano @ 10 & 1 & 1 & 0.876 & 0.983 \\
        \hline
        O4-Mini @ 10 & 1 & 1 & 0.855 & \textbf{0.995} \\
        \hline
    \end{tabular}
    \caption{Results on 70 QA pairs comparing Baseline Non-RAG LLMs with RAG models}
    \label{tab:ragas_results}
\end{table}

\footnotetext[2]{Metrics omitted due to models being non RAG-based, therefore not having a retrieval phase.}

\subsection{Clinical Evaluation Results}
\vspace{5pt}
The clinical evaluation by healthcare professionals provides an important validation of the system's real-world applicability. Table \ref{tab:clinical_evaluation} presents the comparative performance between O4-Mini and GPT-4.1 as assessed by the panel of seven clinicians.\\

\begin{table}[h!]
    \centering
    \small
    \captionsetup{justification=centering}
    \renewcommand{\arraystretch}{1.2}
    \begin{tabular}{|l|c|c|}
        \hline
        \textbf{Metric} & \textbf{O4-Mini} & \textbf{GPT-4.1} \\
        \hline
        Average Accuracy Score & 67.6/70 (96.6\%) & 69.1/70 (98.7\%) \\
        \hline
        Average Count of Complete Answers & 65.6 & 68.3 \\
        \hline
        Average Count of Partial Answers & 4.4 & 1.8 \\
        \hline
        Average Count of Wrong Answers & 0.0 & 0.0 \\
        \hline
        Average Unsafe Responses (per evaluator) & 3.0 & 1.0 \\
        \hline
    \end{tabular}
    \caption{Clinical evaluation results comparing O4-Mini and GPT-4.1 performance as assessed by 7 SMEs on 70 queries}
    \label{tab:clinical_evaluation}
\end{table}

The results demonstrate a clear improvement in the system's performance when utilising the GPT-4.1 model. Not only did the accuracy score increase from 67.6/70 (96.6\%) to 69.1/70 (98.7\%), but more importantly, the safety profile improved significantly. The average number of unsafe responses dropped from 3.0 to 1.0 per evaluator, indicating that GPT-4.1's output was both more complete and safer than that of O4-Mini. The unsafe scores were primarily due to incomplete answers which missed key details e.g. full criteria for cancer screening, and not due to hallucinations. These findings from clinical experts corroborate the automated evaluation results, confirming the system's practical utility for healthcare professionals with the caveat that care should be taken to always refer to the full referenced context and have a human in the loop.\\

\section{Discussion:}
\vspace{5pt}
This study provides one of the first large-scale evaluations of a generative RAG system on a national-level clinical guideline corpus, marking a significant advance in the field. The results confirm the findings of smaller-scale studies, such as those by \cite{ferber2024gpt} and \cite{kresevic2024}, showing clear benefits of implementing RAG architectures over baseline LLMs. The findings demonstrate that this approach not only makes complex medical information more accessible but also significantly enhances the trust in LLMs for clinical queries. The implications of this work are far-reaching, addressing critical challenges in healthcare information retrieval and the safe use of LLMs in healthcare.\\

The primary implication of this research is the demonstration of a reliable and scalable solution to the problem of underutilised clinical guidelines. As noted by \cite{wang_barriers}, the time required for clinicians to manually search through these documents has become a significant barrier to their use. This system directly addresses that challenge by providing precise, contextually grounded answers to queries within seconds (\(\sim\) 5-10 seconds on average). The architecture is also inherently scalable, with the potential to expand beyond the three hundred guidelines indexed here to encompass all guidance offered by NICE as well as guidance from other bodies, such as the various British royal colleges. International guidelines could also be included by leveraging the multilingual capabilities of modern LLMs which would support the greater delivery of evidence-based medicine globally.\\ 

While the current system relies on the NICE clinical guidelines (approximately 300 documents), the architecture is designed to be modular. The vector database containing the NICE guidelines can be readily swapped for a local hospital's specific guidelines or protocols. This flexibility allows the RAG system to be adapted to specific clinical environments, ensuring that the generated responses align with local best practices and formulary or procedural restrictions. This modularity is particularly valuable for NHS trusts and international healthcare systems that may need to incorporate institution-specific protocols alongside national guidelines.\\

The most critical finding of this study is the dramatic reduction in model ``hallucinations". The RAG-enhanced O4-Mini model achieved a faithfulness score of 99.5\%, a 64.7 percentage point increase over its baseline performance. This starkly contrasts with the medically focused Meditron3-8B model, which scored only 43\% on the same metric, underscoring the unreliability of even domain-specific models without the grounding context provided by a RAG architecture.\\

The clinical evaluation comparing O4-Mini and GPT-4.1 further highlights a potential difference in LLM capabilities with regards to reasoning versus non-reasoning models. The reduction in unsafe outputs by two-thirds (from 3.0 to 1.0 per evaluator), with the non-reasoning GPT-4.1 model outperforming the reasoning O4-mini model, represents a crucial finding for the deployment of AI in healthcare. The ability to set a temperature parameter of 0 for non-reasoning models allows for deterministic reproducible outputs, while reasoning models without this feature make it harder to test clinical safety and introduces a layer of unpredictability. This improvement, along with the 98.7\% accuracy score, suggests that non-reasoning models are better at adhering to the provided context and therefore should be prioritised for health domain applications.\\

However, the persistence of \textit{any} unsafe outputs (average of 1.0) underscores that safety risk remains a primary barrier to the adoption of generative AI in clinical settings. This necessitates continued human-in-the-loop verification and robust guardrails before such systems can be used autonomously in patient care. One of the main takeaways is that clinicians who use the system should always refer to the referenced context retrieved to ensure safety and an added note being that only using one set of guidelines may not be enough in clinical practice to be comprehensive.\\

Despite the strong results, this study has limitations that present avenues for future research. The retrieval and completion evaluations were conducted using a synthetically generated dataset of 7901 question-chunk pairs and a curated set of seventy question-answer pairs respectively. While this allowed for extensive testing, these queries may not fully represent the complexity of questions posed by human users in a real-world clinical setting. Cases where required chunks were poorly retrieved were found to arise generally because of a lack of specificity in the query rather than an inherent retrieval failure. Therefore, future work must involve further rigorous testing with healthcare professionals to evaluate the system's real-world usability and performance.\\

A deeper analysis of the completion results reveals limitations in the RAGAs evaluation framework. Firstly, the lower Response Relevancy scores recorded were due to the metric's use of embedding similarity scores for its evaluation. Due to the impact semantic noise and subtle rewording can have on these similarity scores when comparing texts, certain scores were found to be degraded regardless of the response's true relevancy. Secondly, the Faithfulness score of 99.5\% for the O4-Mini model was imperfect due to three anomalies flagged by the RAGAs framework, the most penalised of which is shown in Table \ref{tab:faithfulness_error}. This error was not due to the RAG system itself, but rather a misjudgement by the evaluation framework's LLM judge.\\

\begin{figure}[H]
\centering
\includegraphics[height=8.5cm]{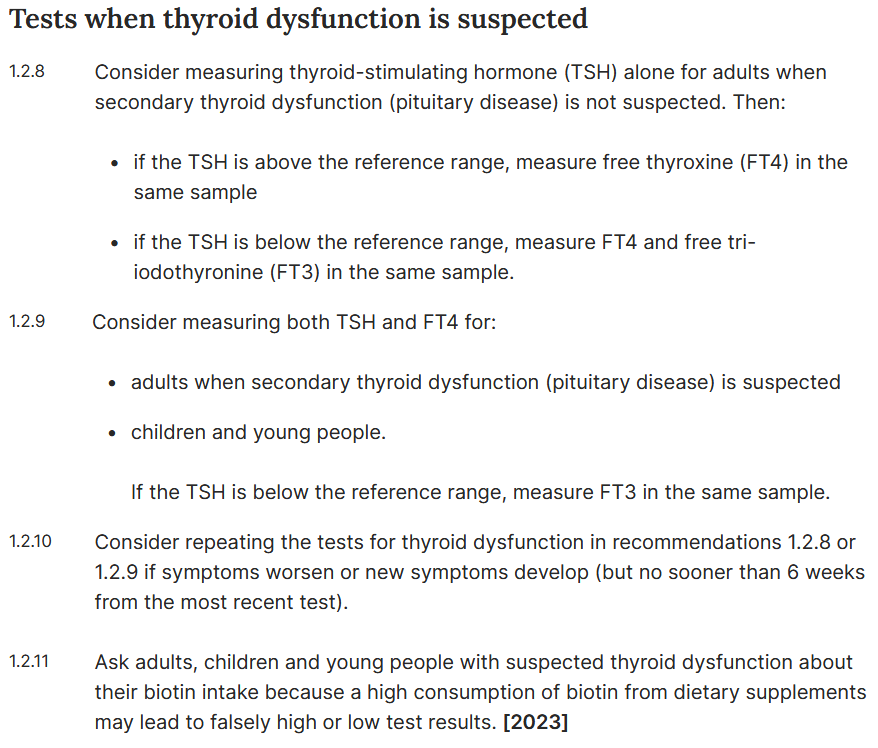}
\caption{Extract of NG145}
\label{fig:ng145screenshot}
\end{figure}

In this case, the model failed to correctly reason that the factually correct claim made in the system's answer was appropriately supported by the context, as shown in Figure \ref{fig:ng145screenshot}. This was due to the guideline formatting, in this case sentence indentation, which led the evaluator LLM to infer that FT3 should only be measured in the same sample as FT4 for children and young people if TSH is below the reference range, and therefore not for adults. This highlights a critical limitation of automated testing and reinforces the need for rigorous real-world trials with healthcare professionals, while also highlighting the impact source document formatting can have within RAG systems.\\

\begin{table}[h!]
\centering
\small
\begin{tabular}{>{\raggedright\arraybackslash}p{0.2\textwidth} >{\raggedright\arraybackslash}p{0.7\textwidth}}
\toprule
\textbf{Question} & What tests should be considered for adults with suspected thyroid dysfunction? \\
\midrule
\textbf{Source} & Guideline NG145 Section 1.2.8-1.2.11 \\
\midrule
\textbf{Answer} & For adults with suspected thyroid dysfunction, consider the following tests:
\begin{itemize}
    \item measuring thyroid-stimulating hormone (TSH) alone when secondary thyroid dysfunction (pituitary disease) is not suspected
        \begin{itemize}
            \itemsep0em
            \item if TSH is above the reference range, measure free thyroxine (FT4) in the same sample
            \item if TSH is below the reference range, measure FT4 and free tri-iodothyronine (FT3) in the same sample
        \end{itemize}
    \item measuring both TSH and FT4 when secondary thyroid dysfunction (pituitary disease) is suspected
        \begin{itemize}
            \itemsep0em
            \item \textcolor{red}{if TSH is below the reference range, measure FT3 in the same sample}
        \end{itemize}
    \item repeating the above tests if symptoms worsen or new symptoms develop (but no sooner than 6 weeks from the most recent test)
    \item asking about biotin intake, because high‐dose biotin supplements may lead to falsely high or low thyroid function test results
\end{itemize} \\
\midrule
\textbf{Original Faithfulness Score} & 0.857 \\
\midrule
\textbf{RAGAs Reasoning} & The context states that for children and young people, if TSH is below the reference range, measure FT3 in the same sample; however, it does not explicitly say to measure FT3 if TSH is below the reference range and secondary thyroid dysfunction is suspected in adults. The statement incorrectly combines conditions not directly supported by the context. \\
\midrule
\textbf{Legend} & \textcolor{red}{Red:} Penalised sentence in the system's answer.\\
\bottomrule
\end{tabular}
\caption{Analysis of Worst O4-Mini Faithfulness Score}
\label{tab:faithfulness_error}
\end{table}

While the conducted testing relied on question-answer pairs curated to require responses grounded to a single guideline at a time, real-world queries often require information to be retrieved from multiple sources at once to be fully addressed. Therefore, further testing must be conducted to evaluate LLMs ability to answer queries necessitating larger information source pools. Additionally, robust testing of the system ability to reject queries with no suitable guideline information was not conducted. To ensure the safety of a future system, this must be addressed in future work.\\

Additionally, the discrepancy between RAGAs results and human evaluation results for safety was due to RAGAs' lack of metric to measure completeness. While the faithfulness metric will detect if generated information is incorrect, it will not detect a lack of required information. This is due to the design of the metric instead of an inherent limitation in LLMs. Therefore further work must be done in designing more comprehensive metrics before LLM-as-a-judge frameworks can fully replace human evaluators, in particular for the medical domain.\\

Another area for development is the exploration of open-source models. The Qwen3-Embedding-0.6B model, while outperformed by some of its closed-source counterparts, could potentially achieve competitive performance if fine-tuned on the specific medical terminology and structure of the NICE guidelines. Additionally, the use of larger open-source counterparts would likely improve results. Furthermore, using open-source models would create a more transparent system which could be hosted on-premises within clinical settings, addressing key data privacy concerns.\\

Beyond its technical performance, the proposed system offers practical strengths in terms of implementation cost-effectiveness and long-term maintainability. At approximately \$0.015 per query, using the highest-performance model combination (Voyage-3-Large, Reranker-2, and O4-Mini), the system is financially viable for deployment. Crucially, its knowledge base supports efficient, incremental updates; new or revised guidelines can be integrated into the vector database without requiring complete, resource-intensive retraining of the entire system.\\

Finally, the transition from a research prototype to a live clinical tool requires careful consideration of risks. The system's integrity is dependent on the accuracy of its source material. The RAG architecture prevents the model from inventing information, but it cannot verify the facts it retrieves. If a NICE guideline contains an error, the system will faithfully present that incorrect information to the user. This vulnerability necessitates a robust process for continuous validation of the underlying knowledge base to ensure safety.\\

\section{Ethical Considerations:}
\vspace{5pt}
Since the system's knowledge base was constructed exclusively from publicly accessible guidelines, meaning no patient data was used, a formal ethical review was not required. However, key ethical considerations must nevertheless be addressed.\\

Firstly, the system's architecture, which leverages external third-party LLM APIs, presents a potential privacy problem. While their use is acceptable for non-sensitive public data, it presents a critical risk for clinical deployment. Should a user input sensitive patient information, transmitting such data to external servers would violate data protection frameworks like the GDPR and HIPAA, unless cloud providers have signed compliance agreements. Therefore, an essential mitigation strategy for any transition to a live clinical environment is architectural modification. The recommended approach would therefore be the on-premises hosting of open-source models within a clinical network.\\

Finally, algorithmic accountability, specifically model hallucination, was a central ethical consideration in this paper. While the system’s high faithfulness metrics confirm the success of RAG at minimising hallucination, a non-zero risk of generating content that deviates from the source context persists. This risk mandates that the system must function strictly as an adjunctive clinical tool. The ethical and safe implementation of such a system would rely on human-in-the-loop oversight, where the clinical judgement of a healthcare professional remains the final arbiter in any decision-making process.\\

\section{Conclusion:}
\vspace{5pt}
This project successfully developed and evaluated a RAG system designed to make the UK's NICE guidelines more accessible using LLMs. By addressing the significant challenge of navigating these extensive and complex documents, the system provides a tool for healthcare professionals to obtain rapid, evidence-based answers to clinical questions while offering a flexible, modular framework that can be adapted to any clinical guideline corpus.\\

The evaluation demonstrated the system's high performance at every stage. The retrieval architecture proved exceptionally effective, achieving a Mean Reciprocal Rank (MRR) of 0.814 and a near-perfect Recall@10 of 99.1\%, ensuring that relevant information was consistently identified from the vast knowledge base of guidelines. The most critical contribution, however, was observed in the generation phase, where the RAG methodology dramatically enhanced the reliability of the LLM outputs. The system achieved a Faithfulness score of 0.995 and a perfect Context Precision score of 1.0, confirming its ability to generate answers that are strictly grounded in the source material. This stands in stark contrast to non-RAG models, which exhibited a high propensity for hallucination, scoring as low as 0.348 on the same Faithfulness metric. This was further validated by clinical experts, where the GPT-4.1 model achieved a 98.7\% accuracy score, though the persistence of unsafe outputs reinforces the need for human oversight.\\

In conclusion, this research validates RAG as a robust, scalable, and effective strategy for deploying LLMs safely within a clinical information setting. By successfully mitigating the risk of information fabrication, this work presents a significant step toward the responsible integration of generative AI into healthcare. The developed system offers a practical and cost-effective solution that can reduce the time required to access critical guidance, thereby supporting busy clinicians and enhancing patient care. Future work should build on these findings, moving towards human-centric evaluations to prepare the system for deployment in a real-world clinical environment.\\

\section{Acknowledgment:}
\vspace{5pt}
The authors gratefully acknowledge the support provided by Iain Moir and Angus Leitch at the National Institute of Health and Care Excellence (NICE) during the course of this project.

\section{Data Availability}
The clinical guidelines analysed in this study are publicly available via the NICE API. The specific subset of guidelines used in this study was acquired on July 16, 2025. The synthetic query datasets generated during the study are available at github.com/matthewlewis123/A-NICE-RAG.

\section{Code Availability}
The code used for the Retrieval-Augmented Generation system, including the pre-processing pipeline and evaluation framework, is available at github.com/matthewlewis123/A-NICE-RAG.

\newpage
\bibliography{references}   

\newpage
\appendix
\section*{Appendix}
\vspace{5pt}
\section{Excluded Sections for Query Generation}
\vspace{5pt}
The following sections were excluded during the query generation process:\\

\begin{itemize}
    \item About this quality standard
    \item Appendix
    \item Appraisal committee members
    \item Committee members
    \item Diagnostics advisory committee members and NICE project team
    \item Endorsing organisation
    \item Evaluation committee members and NICE project team
    \item Finding more information and committee details
    \item Putting this guideline into practice
    \item Sources of evidence
    \item Supporting organisations
    \item Update information
\end{itemize}
\vspace{5pt}
\section{BM25 Hyperparameters}
\vspace{5pt}
\begin{table}[h!]
\centering
\begin{tabular}{|l|l|}
\hline
\textbf{Hyperparameter} & \textbf{Value} \\ \hline
Tokenization & Lemmatised \\ \hline
$k_1$ & 1.7 \\ \hline
$b$ & 0.83 \\ \hline
$\epsilon$ & 0.05 \\ \hline
\end{tabular}
\caption{Hyperparameters for BM25 Indexing}
\label{tab:bm25_hyperparameters}
\end{table}
\vspace{5pt}

\section{Hybrid-Search Hyperparameters}
\vspace{5pt}
\begin{table}[h!]
\small
\centering
\begin{tabularx}{\textwidth}{|l|>{\centering\arraybackslash}X|>{\centering\arraybackslash}X|c|}
\hline
\textbf{Model Combination} 
& \textbf{Component 1 Weight ($w_{1}$)} 
& \textbf{Component 2 Weight ($w_{2}$)} 
& \textbf{Constant ($k$)} \\ \hline
Voyage-3-Large + BM25 & 5 & 1 & 40 \\ \hline
Voyage-3-Large + Text-Embedding-3-Large & 2 & 1 & 40 \\ \hline
\end{tabularx}
\caption{WRRF Weight Values for Retrieval Stage Model Combinations}
\label{tab:wrrf_weights}
\end{table}

\section{Theoretical Cost Calculation}
\vspace{5pt}
\begin{table}[H]
\centering
\small
\begin{tabular}{|l|p{5.5cm}|r|r|r|}
\hline
\textbf{Component} & \textbf{Token Calculation Details} & \textbf{Tokens} & \textbf{\makecell{Price per 1M \\ Tokens (\$)}} & \textbf{Cost (\$)} \\ \hline
Embedding & Query & 100 & 0.18 & 0.000018 \\ \hline
Reranking & 15 chunks * (100 Query Tokens + 500 Chunk Tokens) & 9,000 & 0.05 & 0.00045 \\ \hline
LLM Input & 650 (System and User Prompt) + 100 (Query) + 10 * 500 (Context) & 5,750 & 2.00 & 0.0115 \\ \hline
LLM Output & Assumed answer length & 500 & 8.00 & 0.004 \\ 
\hline
\textbf{Total} & & \textbf{15,350} & - & \textbf{0.015968} \\ \hline
\end{tabular}
\caption{Theoretical Cost Calculation for a Single Query}
\label{tab:cost_calculation}
\end{table}

\section{RAGAs Faithfulness Failures O4-Mini}
\vspace{5pt}
\begin{table}[h!]
\centering
\footnotesize
\begin{tabular}{>{\raggedright\arraybackslash}p{0.2\textwidth} >{\raggedright\arraybackslash}p{0.7\textwidth}}
\toprule
\textbf{Component} & \textbf{Content} \\
\midrule
\textbf{Question} & What are the recommended treatment options for neurogenic stress urinary incontinence? \\
\midrule
\textbf{Source} & Guideline CG148 Section 1.4 \\
\midrule
\textbf{Answer} & 
Recommended treatment options for neurogenic stress urinary incontinence include:
\begin{itemize}
    \itemsep0em 
    \item Pelvic floor muscle training for people with neurogenic lower urinary tract dysfunction where voluntary pelvic floor contraction is preserved. Consider combining this with
        \begin{itemize}
            \itemsep0em
            \item biofeedback
            \item electrical stimulation of the pelvic floor
        \end{itemize}
    \item Autologous fascial sling surgery
    \item Artificial urinary sphincter insertion for people in whom an autologous fascial sling is less likely to control incontinence. \textcolor{red}{Before insertion:}
        \begin{itemize}
            \itemsep0em
            \item discuss the risks, need for possible repeat operations and alternative procedures
            \item ensure the bladder has adequate low-pressure storage capacity
            \item \textcolor{red}{plan for upper urinary tract monitoring (for example, annual ultrasound scans)}
        \end{itemize}
\end{itemize} \\
\midrule
\textbf{Original Faithfulness Score} & 0.875 \\
\midrule
\textbf{RAGAs Reasoning} & The context states to monitor the upper urinary tract after artificial urinary sphincter surgery, for example using annual ultrasound scans, but does not specify planning this before insertion.\\
\midrule
\textbf{Legend} & \textcolor{red}{Red:} Penalised sentence in the system's answer.\\
\bottomrule
\end{tabular}
\caption{Analysis of Second Worst O4-Mini Faithfulness Score}
\label{tab:faithfulness_error2}
\end{table}

\begin{figure}[h]
\centering
\includegraphics[height=8cm]{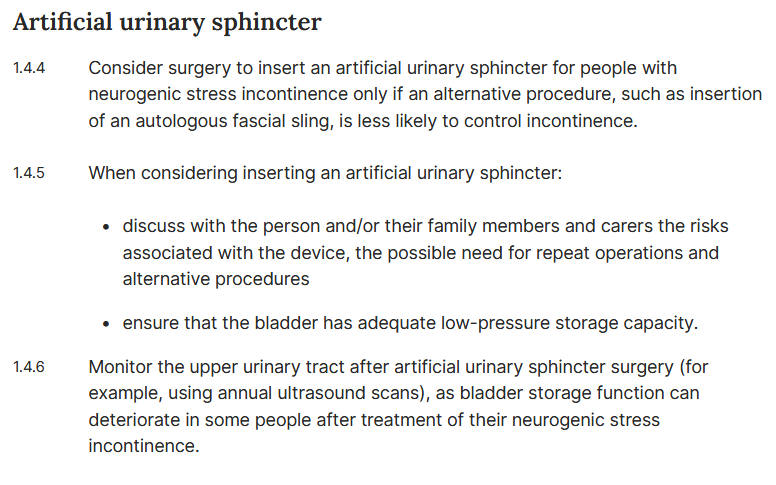}
\caption{Extract of CG148}
\label{fig:cg148screenshot}
\end{figure}

\begin{table}[h!]
\centering
\footnotesize
\begin{tabular}{>{\raggedright\arraybackslash}p{0.2\textwidth} >{\raggedright\arraybackslash}p{0.7\textwidth}}
\toprule
\textbf{Component} & \textbf{Content} \\
\midrule
\textbf{Question} & When and how should mirror therapy be used as part of stroke rehabilitation for people with upper limb weakness? \\
\midrule
\textbf{Source} & Guideline NG236 Section 1.13.30–1.13.31 \\
\midrule
\textbf{Answer} & 
According to NICE guideline NG236 (recommendations 1.13.30–1.13.31), mirror therapy for upper limb weakness after stroke should be used as follows:
\begin{itemize}
    \item[\textbf{Consideration}]
        \begin{itemize}
            \itemsep0em
            \item Consider mirror therapy as an adjunct to the person’s existing rehabilitation programme.
            \item \textcolor{red}{Offer} it to people with upper (or lower) limb muscle weakness after stroke.
        \end{itemize}
    \item[\textbf{Timing}]
        \begin{itemize}
            \itemsep0em
            \item Start mirror therapy within the first 6 months after stroke (ideally within 1 month).
        \end{itemize}
    \item[\textbf{Session parameters}]
        \begin{itemize}
            \itemsep0em
            \item Sessions should be around 30 minutes long.
            \item Hold sessions at least 5 times per week for 4 weeks.
        \end{itemize}
    \item[\textbf{Supervision}]
        \begin{itemize}
            \itemsep0em
            \item Initially supervise mirror therapy (for example, by a physiotherapist or occupational therapist), focusing on how it works and what to expect.
            \item Continue supervision as needed (for example, for people with cognitive difficulties).
            \item Once trained, many people can continue therapy unsupervised in hospital or at home (with a loaned mirror).
        \end{itemize}
\end{itemize}
For details of the evidence and committee discussion, see the \href{https://www.nice.org.uk/guidance/ng236/evidence/q-mirror-therapy-pdf-474910116413}{evidence review Q: mirror therapy}. \\
\midrule
\textbf{Original Faithfulness Score} & 0.9 \\
\midrule
\textbf{RAGAs Reasoning} & The context uses the term `consider mirror therapy' rather than `offer', so it does not directly support that mirror therapy should be offered to all such people.\\
\midrule
\textbf{Legend} & \textcolor{red}{Red:} Penalised sentence in the system's answer.\\
\bottomrule
\end{tabular}
\caption{Analysis of Third Worst O4-Mini Faithfulness Score}
\label{tab:faithfulness_error3}
\end{table}

\clearpage

\begin{figure}[h]
\centering
\includegraphics[height=3cm]{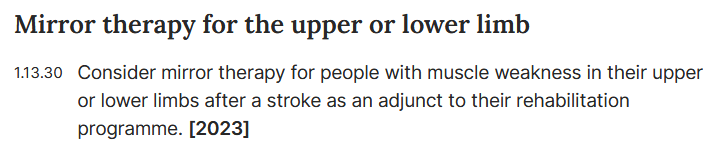}
\caption{Extract of NG236}
\label{fig:ng236screenshot}
\end{figure}

\section{Changes to RAGAs System Prompts}
\vspace{5pt}
This appendix details the modifications made to the default system prompts of the RAGAs framework metrics. The prompts were tailored to enhance their performance and relevance within the medical domain, specifically for querying NICE clinical guidelines. Each table below presents a side-by-side comparison of the original prompt and the modified version used in this study, followed by tables containing the full original and modified examples.

\renewcommand{\arraystretch}{1.5}

\subsection{Faithfulness Metric}
\vspace{5pt}
\begin{longtable}{@{}p{0.45\textwidth} p{0.45\textwidth}@{}}
\caption{Modifications to the \texttt{faithfulness} Metric Prompts}
\label{tab:faithfulness_prompts} \\
\toprule
\textbf{Original Prompt} & \textbf{Modified Prompt} \\
\midrule
\endfirsthead
\caption*{(Continued) Modifications to the \texttt{faithfulness} Metric Prompts} \\
\toprule
\textbf{Original Prompt} & \textbf{Modified Prompt} \\
\midrule
\endhead
\bottomrule
\endfoot
\multicolumn{2}{c}{\textit{Prompt 1: Statement Generation}} \\
\midrule
Given a question and an answer, analyze the complexity of each sentence in the answer. Break down each sentence into one or more fully understandable statements. Ensure that no pronouns are used in any statement. Format the outputs in JSON. & Given a medical question and an answer, break down the answer into fully understandable statements relevant to clinical guidelines. Ensure that no pronouns are used in any statement. Format the outputs in JSON. \\
\midrule
\multicolumn{2}{c}{\textit{Prompt 2: Statement Verification}} \\
\midrule
Your task is to judge the faithfulness of a series of statements based on a given context. For each statement you must return verdict as 1 if the statement can be directly inferred based on the context or 0 if the statement can not be directly inferred based on the context. & Your task is to judge the faithfulness of a series of statements based on a given medical context. For each statement, return verdict as 1 if the statement can be directly inferred from the context or 0 if the statement cannot be directly inferred from the context. Output JSON with reasoning. \\
\end{longtable}

\begin{longtable}{@{}p{\textwidth}@{}}
\caption{Original Examples for \texttt{faithfulness} Metric} \\
\toprule
\textbf{Statement Generation Example} \\
\midrule
\endfirsthead
\bottomrule
\endfoot
\textbf{Input}: \\
\texttt{question}: ``Who was Albert Einstein and what is he best known for?'' \\
\texttt{answer}: ``He was a German-born theoretical physicist, widely acknowledged to be one of the greatest and most influential physicists of all time. He was best known for developing the theory of relativity, he also made important contributions to the development of the theory of quantum mechanics.'' \\
\midrule
\textbf{Output}: \\
\texttt{statements}: [
    ``Albert Einstein was a German-born theoretical physicist.'',
    ``Albert Einstein is recognized as one of the greatest and most influential physicists of all time.'',
    ``Albert Einstein was best known for developing the theory of relativity.'',
    ``Albert Einstein also made important contributions to the development of the theory of quantum mechanics.''
] \\
\vspace{0.5cm}
\textbf{Statement Verification Example} \\
\midrule
\textbf{Input}: \\
\texttt{context}: ``John is a student at XYZ University. He is pursuing a degree in Computer Science. He is enrolled in several courses this semester, including Data Structures, Algorithms, and Database Management. John is a diligent student and spends a significant amount of time studying and completing assignments. He often stays late in the library to work on his projects.'' \\
\texttt{statements}: [
    ``John is majoring in Biology.'',
    ``John is taking a course on Artificial Intelligence.'',
    ``John is a dedicated student.'',
    ``John has a part-time job.''
] \\
\midrule
\textbf{Output}: \\
\texttt{statements}: [ \\
    \{
        ``statement'': ``John is majoring in Biology.'',
        ``reason'': ``John's major is explicitly mentioned as Computer Science. There is no information suggesting he is majoring in Biology.'',
        ``verdict'': 0
    \}, \\
    \{
        ``statement'': ``John is a dedicated student.'',
        ``reason'': ``The context states that he spends a significant amount of time studying and completing assignments. Additionally, it mentions that he often stays late in the library to work on his projects, which implies dedication.'',
        ``verdict'': 1
    \}
] \\
\end{longtable}

\begin{longtable}{@{}p{\textwidth}@{}}
\caption{Modified Examples for \texttt{faithfulness} Metric} \\
\toprule
\textbf{Statement Generation Example} \\
\midrule
\endfirsthead
\bottomrule
\endfoot
\textbf{Input}: \\
\texttt{question}: ``What are risk factors for ADHD?'' \\
\texttt{answer}: ``Children born preterm are at significantly higher risk for developing attention deficit hyperactivity disorder (ADHD) compared to those born at term. Girls may be under-recognised for ADHD due to differences in symptom presentation and referral patterns. Family history, especially having a close relative diagnosed with ADHD, is considered a major risk factor for developing the disorder.'' \\
\midrule
\textbf{Output}: \\
\texttt{statements}: [
    ``Children born preterm are at higher risk for ADHD.'',
    ``Girls may be under-recognised for ADHD.'',
    ``Family history is a major risk factor for ADHD.''
] \\
\vspace{0.5cm}
\textbf{Statement Verification Example} \\
\midrule
\textbf{Input}: \\
\texttt{context}: ``Consider antihypertensive drug treatment in addition to lifestyle advice for people aged over 80 with stage 1 hypertension if their clinic blood pressure is over 150/90 mmHg. Use clinical judgement for people with frailty or multimorbidity.'' \\
\texttt{statements}: [
    ``Antihypertensive drug treatment should be considered for people aged over 80 with stage 1 hypertension if their clinic blood pressure is over 150/90 mmHg.'',
    ``Antihypertensive drug treatment should be considered for people aged over 80 with stage 1 hypertension regardless of blood pressure.''
] \\
\midrule
\textbf{Output}: \\
\texttt{statements}: [ \\
    \{
        ``statement'': ``Antihypertensive drug treatment should be considered for people aged over 80 with stage 1 hypertension if their clinic blood pressure is over 150/90 mmHg.'',
        ``reason'': ``This statement is directly supported by the context.'',
        ``verdict'': 1
    \}, \\
    \{
        ``statement'': ``Antihypertensive drug treatment should be considered for people aged over 80 with stage 1 hypertension regardless of blood pressure.'',
        ``reason'': ``The context specifies that treatment should be considered only if clinic blood pressure is over 150/90 mmHg.'',
        ``verdict'': 0
    \}
] \\
\end{longtable}

\vspace{0.5cm}

\subsection{Answer Relevancy Metric}
\vspace{5pt}
\begin{longtable}{@{}p{0.45\textwidth} p{0.45\textwidth}@{}}
\caption{Modification to the \texttt{answer\_relevancy} Metric Prompt} \\
\toprule
\textbf{Original Prompt} & \textbf{Modified Prompt} \\
\midrule
\endfirsthead
\bottomrule
\endfoot
Generate a question for the given answer and Identify if answer is noncommittal. Give noncommittal as 1 if the answer is noncommittal and 0 if the answer is committal. A noncommittal answer is one that is evasive, vague, or ambiguous. For example, ``I don't know'' or ``I'm not sure'' are noncommittal answers. & Given an answer based on NICE clinical guidelines, generate a relevant question that this answer addresses. Also, identify if the answer is noncommittal. Set `noncommittal' to 1 if the answer is vague, evasive, or ambiguous (e.g., ``No relevant NICE guidelines were found''), and 0 if the answer is committal. \\
\end{longtable}
\clearpage
\begin{longtable}{@{}p{\textwidth}@{}}
\caption{Original Examples for \texttt{answer\_relevancy} Metric} \\
\toprule
\endfirsthead
\bottomrule
\endfoot
\textbf{Example 1 Input}: \\
\texttt{response}: ``Albert Einstein was born in Germany.'' \\
\midrule
\textbf{Example 1 Output}: \\
\texttt{question}: ``Where was Albert Einstein born?'' \\
\texttt{noncommittal}: 0 \\
\midrule
\textbf{Example 2 Input}: \\
\texttt{response}: ``I don't know about the groundbreaking feature of the smartphone invented in 2023 as am unaware of information beyond 2022.'' \\
\midrule
\textbf{Example 2 Output}: \\
\texttt{question}: ``What was the groundbreaking feature of the smartphone invented in 2023?'' \\
\texttt{noncommittal}: 1 \\
\end{longtable}
\vspace{0.5cm}
\begin{longtable}{@{}p{\textwidth}@{}}
\caption{Modified Examples for \texttt{answer\_relevancy} Metric} \\
\toprule
\endfirsthead
\bottomrule
\endfoot
\textbf{Example 1 Input}: \\
\texttt{response}: ``Offer methylphenidate (either short or long acting) as the first line pharmacological treatment for children aged 5 years and over and young people with ADHD.'' \\
\midrule
\textbf{Example 1 Output}: \\
\texttt{question}: ``What is the first-line pharmacological treatment for children aged 5 years and over with ADHD?'' \\
\texttt{noncommittal}: 0 \\
\midrule
\textbf{Example 2 Input}: \\
\texttt{response}: ``No relevant NICE guidelines were found to answer your question.'' \\
\midrule
\textbf{Example 2 Output}: \\
\texttt{question}: ``What is the first-line treatment for hypertension?'' \\
\texttt{noncommittal}: 1 \\
\end{longtable}
\newpage

\subsection{Context Precision Metric}
\vspace{5pt}
\begin{longtable}{@{}p{0.45\textwidth} p{0.45\textwidth}@{}}
\caption{Modification to the \texttt{context\_precision} Metric Prompt} \\
\toprule
\textbf{Original Prompt} & \textbf{Modified Prompt} \\
\midrule
\endfirsthead
\bottomrule
\endfoot
Given question, answer and context verify if the context was useful in arriving at the given answer. Give verdict as ``1'' if useful and ``0'' if not with json output. & Given question, answer and context verify if the context was useful in arriving at the given answer. Give verdict as ``1'' if useful and ``0'' if not with json output. \\
\end{longtable}

\begin{longtable}{@{}p{\textwidth}@{}}
\caption{Original Examples for \texttt{context\_precision} Metric} \\
\toprule
\endfirsthead
\bottomrule
\endfoot
\textbf{Example 1 Input}: \\
\texttt{question}: ``What can you tell me about Albert Einstein?'' \\
\texttt{context}: ``Albert Einstein (14 March 1879 – 18 April 1955) was a German-born theoretical physicist, widely held to be one of the greatest and most influential scientists of all time...'' \\
\texttt{answer}: ``Albert Einstein, born on 14 March 1879, was a German-born theoretical physicist, widely held to be one of the greatest and most influential scientists of all time. He received the 1921 Nobel Prize in Physics for his services to theoretical physics.'' \\
\midrule
\textbf{Example 1 Output}: \\
\texttt{reason}: ``The provided context was indeed useful in arriving at the given answer. The context includes key information about Albert Einstein's life and contributions, which are reflected in the answer.'' \\
\texttt{verdict}: 1 \\
\midrule
\textbf{Example 2 Input}: \\
\texttt{question}: ``What is the tallest mountain in the world?'' \\
\texttt{context}: ``The Andes is the longest continental mountain range in the world, located in South America...'' \\
\texttt{answer}: ``Mount Everest.'' \\
\midrule
\textbf{Example 2 Output}: \\
\texttt{reason}: ``the provided context discusses the Andes mountain range, which, while impressive, does not include Mount Everest or directly relate to the question about the world's tallest mountain.'' \\
\texttt{verdict}: 0 \\
\end{longtable}

\begin{longtable}{@{}p{\textwidth}@{}}
\caption{Modified Examples for \texttt{context\_precision} Metric} \\
\toprule
\endfirsthead
\bottomrule
\endfoot
\textbf{Example 1 Input}: \\
\texttt{question}: ``How often should children with type 1 diabetes be screened for thyroid disease?'' \\
\texttt{context}: ``Offer children and young people with type 1 diabetes monitoring for: • thyroid disease, at diagnosis and then annually until transfer to adult services...'' \\
\texttt{answer}: ``Thyroid disease should be screened at diagnosis and then annually until transfer to adult services.'' \\
\midrule
\textbf{Example 1 Output}: \\
\texttt{reason}: ``The context directly states the recommended screening frequency for thyroid disease, which matches the frequency in the response provided.'' \\
\texttt{verdict}: 1 \\
\midrule
\textbf{Example 2 Input}: \\
\texttt{question}: ``Is atomoxetine recommended as the first line pharmacological treatment for children aged 5 years and over with ADHD?'' \\
\texttt{context}: ``Offer methylphenidate (either short or long acting) as the first line pharmacological treatment for children aged 5 years and over and young people with ADHD...'' \\
\texttt{answer}: ``Atomoxetine is recommended as the first line pharmacological treatment for children aged 5 years and over with ADHD.'' \\
\midrule
\textbf{Example 2 Output}: \\
\texttt{reason}: ``The context recommends methylphenidate as the first line treatment for children aged 5 years and over with ADHD, not atomoxetine. Therefore, the context does not support the answer provided.'' \\
\texttt{verdict}: 0 \\
\end{longtable}

\clearpage

\subsection{Context Recall Metric}
\vspace{5pt}
\begin{longtable}{@{}p{0.45\textwidth} p{0.45\textwidth}@{}}
\caption{Modification to the \texttt{context\_recall} Metric Prompt} \\
\toprule
\textbf{Original Prompt} & \textbf{Modified Prompt} \\
\midrule
\endfirsthead
\bottomrule
\endfoot
Given a context, and an answer, analyze each sentence in the answer and classify if the sentence can be attributed to the given context or not. Use only `Yes' (1) or `No' (0) as a binary classification. Output json with reason. & Given a context, and an answer, analyze each sentence in the answer and classify if the sentence can be attributed to the given context or not. Use only `Yes' (1) or `No' (0) as a binary classification. Output json with reason. \\
\end{longtable}

\begin{longtable}{@{}p{\textwidth}@{}}
\caption{Original Examples for \texttt{context\_recall} Metric} \\
\toprule
\endfirsthead
\bottomrule
\endfoot
\textbf{Input}: \\
\texttt{question}: ``What can you tell me about Albert Einstein?'' \\
\texttt{context}: ``Albert Einstein (14 March 1879 - 18 April 1955) was a German-born theoretical physicist...'' \\
\texttt{answer}: ``Albert Einstein born in 14 March 1879 was German-born theoretical physicist, widely held to be one of the greatest and most influential scientists of all time. He received the 1921 Nobel Prize in Physics for his services to theoretical physics. He published 4 papers in 1905. Einstein moved to Switzerland in 1895'' \\
\midrule
\textbf{Output}: \\
\texttt{classifications}: [ \\
    \{
        ``statement'': ``Albert Einstein, born on 14 March 1879, was a German-born theoretical physicist, widely held to be one of the greatest and most influential scientists of all time.'',
        ``reason'': ``The date of birth of Einstein is mentioned clearly in the context.'',
        ``attributed'': 1
    \}, \\
    \{
        ``statement'': ``He published 4 papers in 1905.'',
        ``reason'': ``There is no mention about papers he wrote in the given context.'',
        ``attributed'': 0
    \}
] \\
\end{longtable}

\begin{longtable}{@{}p{\textwidth}@{}}
\caption{Modified Examples for \texttt{context\_recall} Metric} \\
\toprule
\endfirsthead
\bottomrule
\endfoot
\textbf{Input}: \\
\texttt{question}: ``What are risk factors for ADHD?'' \\
\texttt{context}: ``People born preterm may have increased prevalence of ADHD compared with the general population. ADHD is thought to be under-recognised in girls and women. Universal screening for ADHD should not be undertaken in schools.'' \\
\texttt{answer}: ``Children born preterm are at higher risk for ADHD. Girls may be under-recognised for ADHD. Family history is a major risk factor for ADHD.'' \\
\midrule
\textbf{Output}: \\
\texttt{classifications}: [ \\
    \{
        ``statement'': ``Children born preterm are at higher risk for ADHD.'',
        ``reason'': ``This statement directly matches the first sentence.'',
        ``attributed'': 1
    \}, \\
    \{
        ``statement'': ``Family history is a major risk factor for ADHD.'',
        ``reason'': ``There is no mention of family history as a risk factor in any sentence in the context.'',
        ``attributed'': 0
    \}
] \\
\end{longtable}

\end{document}